\documentclass[journal]{IEEEtran}

\ifCLASSINFOpdf
\else
\fi

\usepackage[pagebackref=true,breaklinks=true,letterpaper=true,colorlinks,bookmarks=false]{hyperref}

\usepackage[pdftex]{graphicx}
\usepackage{bm}
\usepackage{amsmath}
\usepackage{amssymb}
\usepackage{algorithm}
\usepackage{algorithmic}
\usepackage{mathrsfs}
\usepackage{enumitem} 
\usepackage{multirow}
\usepackage{tabularx}
\usepackage{adjustbox}
\usepackage{makecell}

\graphicspath{image}

\def\eg{\emph{e.g.}}
\def\etal{\emph{et al.}}
\def\ie{\emph{i.e.}}

\def\f{\mathbf{f}}
\def\v{\mathbf{v}}
\def\hatf{\hat{\mathbf{f}}}

\begin{document}
\title{Learning to Anticipate Egocentric Actions by Imagination}

\author{Yu Wu, Linchao Zhu, Xiaohan Wang, Yi Yang, Fei Wu
\thanks{
(Corresponding Author: Yi Yang)

Y. Wu, L. Zhu, X. Wang, and Y. Yang are with the Australian Artificial Intelligence Institute, University of Technology Sydney, Ultimo 2007, NSW, Australia. (E-mail: yu.wu-3@student.uts.edu.au; linchao.zhu@uts.edu.au; xiaohan.wang-3@student.uts.edu.au; yi.yang@uts.edu.au).

F. Wu is with Zhejiang University, Hangzhou 310027, China (Email:wufei@cs.zju.edu.cn).
}
}

\markboth{IEEE TRANSACTIONS ON IMAGE PROCESSING, VOL. 30, 2021}%
{Shell \MakeLowercase{\textit{et al.}}: Bare Demo of IEEEtran.cls for IEEE Journals}

\maketitle

\begin{abstract}
Anticipating actions before they are executed is crucial for a wide range of practical applications, including autonomous driving and robotics. In this paper, we study the egocentric action anticipation task, which predicts future action seconds before it is performed for egocentric videos. Previous approaches focus on summarizing the observed content and directly predicting future action based on past observations. We believe it would benefit the action anticipation if we could mine some cues to compensate for the missing information of the unobserved frames. We then propose to decompose the action anticipation into a series of future feature predictions. We imagine how the visual feature changes in the near future and then predicts future action labels based on these imagined representations.
Differently, our ImagineRNN is optimized in a contrastive learning way instead of feature regression. We utilize a proxy task to train the ImagineRNN, \ie, selecting the correct future states from distractors. We further improve ImagineRNN by residual anticipation, \ie, changing its target to predicting the feature difference of adjacent frames instead of the frame content. This promotes the network to focus on our target, \ie, the future action, as the difference between adjacent frame features is more important for forecasting the future. Extensive experiments on two large-scale egocentric action datasets validate the effectiveness of our method.
Our method significantly outperforms previous methods on both the seen test set and the unseen test set of the EPIC Kitchens Action Anticipation Challenge.
\end{abstract}


\IEEEpeerreviewmaketitle

\section{Introduction}
 \IEEEPARstart{A}{nticipating} the near future is a natural task that has drawn increasing research attention~\cite{chen2018part,li2019way}. It has a wide range of applications in the intelligent systems when it needs to react before an action gets executed. For instance, it is critical to anticipate if a car would stop or a pedestrian would cross the road in the autonomous driving task. The prediction is supposed to be seconds before the action is actually taken place, so that the autonomous vehicle could have time to react to avoid an accident. 
Under these circumstances, recent works are proposed to predict activities a few seconds in the future~\cite{miech2019leveraging,furnari2019rulstm}, which is practical for real-world applications.

In this paper, we focus on the problem of egocentric action anticipation defined in~\cite{furnari2019rulstm}.
Egocentric (First Person Vision) videos~\cite{Damen2018EPICKITCHENS,8489917} offers an interesting scenario to study the action anticipation problem.
Given an egocentric video sequence denoted as observed video, we aim to predict the future action that happens after a time period of $T$ seconds, whereas the time $T$ is known as the anticipation time.

\begin{figure}[t!]
    \begin{center}
        \includegraphics[width=\linewidth]{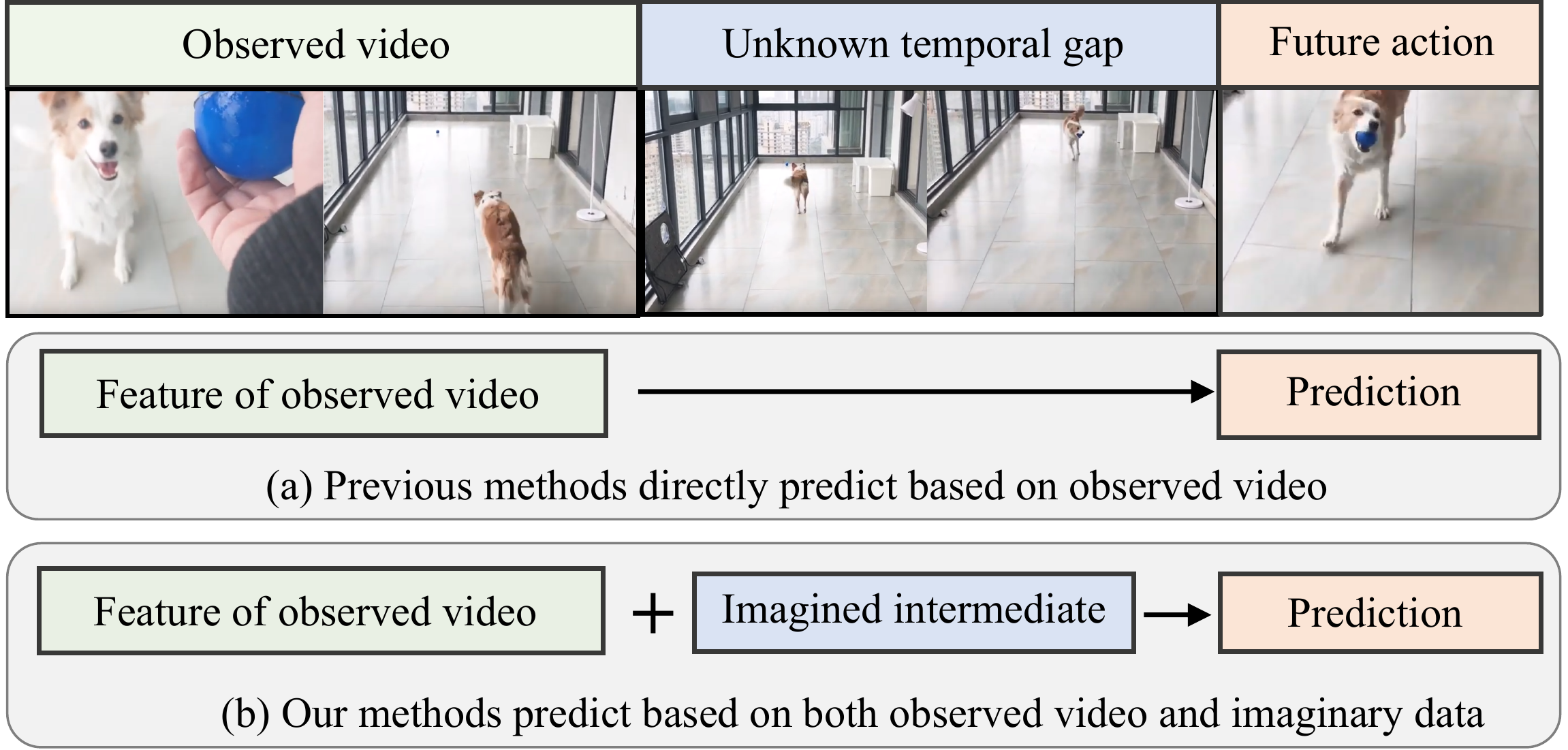}
    \end{center}
    \caption{In the action anticipation task, the model needs to predict the future action that happens $T$ seconds later.
    Predicting the intermediate future features~\cite{liu2018future,luc2018predicting,shi2018action,gammulle2019predicting,rodriguez2018action} would benefit the action anticipation task.
    Our study focuses on how to learn a better imagined intermediate feature.
    }
\label{fig:motivation2}
\end{figure}

Most previous approaches~\cite{miech2019leveraging,furnari2019rulstm} focus on summarizing the past observed frames, and then directly predict the future action that takes place $T$ seconds later. 
These methods overlook the temporal gap between the past observations and the future action that is supposed to be predicted. 
However, frames in this temporal period are closer to the future, thus containing more useful evidence for the next action.
If we could mine cues to compensate for missing information of unobserved frames, it would be easier for anticipation models to predict the future.

In this paper, we propose to tackle this issue by imagining the near future.
First, we decompose the long-time action anticipation into a series of future feature predictions.
We imagine how the visual feature changes in the very near future and then predict the future action labels based on these imagined representations.
Specifically, we design the ImagineRNN to predict the next visual representation based on past observations in a step-wise manner. 
Since our target is to predict the future action, it is unnecessary to waste model capacity on resolving the stochasticity of frame appearance changes due to camera motion and shadows in egocentric videos.
Thus in ImagineRNN, we only generate the visual representation instead of raw pixels.
The final anticipation is built on both the observed content and visual representation that we imagined within the anticipation time $T$.  

Recently, some works~\cite{liu2018future,luc2018predicting,shi2018action,gammulle2019predicting,rodriguez2018action} also propose to generate intermediate future frames or future content features using RNN or GAN architectures. Most of these works use regression loss functions (\eg, $l_2$ loss or cosine loss) or discriminator (justifying real or fake) to optimize their generator model.
However, these optimization methods are too deterministic in training the generator model. 
There are only positive targets in these loss functions, leading to biased or sub-optimal optimization on the predicted future features. 
In addition, since actions are changed very quickly in egocentric videos, the predicted future states should be distinguishable in time sequences.
Optimization with only positive targets would overlook the state changes in the future time period.

Our ImagineRNN differs from existing works in two aspects. 
First is that our ImagineRNN is optimized in the contrastive learning manner instead of feature regression.
We propose a proxy task to train the ImagineRNN by selecting the correct future states from distractors.
For the predicted future feature, we first build a set of candidates containing both the positive target (the ground truth future feature) and negative distractors (features from other time steps).
Then we encourage the model to learn to identify the correct future state from candidates given the observed context.
In this way, our ImagineRNN could essentially learn the change of future features.
We found the new optimization method significantly improves the generalisability on the unseen test set.

Second, we further improve ImagineRNN by residual anticipation, \ie, changing its target to predicting the feature difference of adjacent frames, instead of the entire frame feature.
Different from~\cite{gammulle2019predicting,rodriguez2018action} that predict the entire optical flow frames or dynamic image, we only predict the feature changes between adjacent frames.
The motivation is in three-folds.    
First, the difference between adjacent frame features is more important for forecasting the future. Predicting the video difference promotes the network to focus on the change of intermediate features, leading to better results on the future action anticipation. 
Second, it reduces the load of the ImagineRNN and thus saves the model capacity. In this way, the information the ImagineRNN has to predict is minimized, while the unchanged feature channels are directly carried forward.
Third, the unchanged content plays a role of shortcut connection, avoiding the noise accumulation and the gradient vanishing. 
To the best of our knowledge, we are the first to forecast the difference of frames in generating future features.

We conduct extensive experiments on two large-scale egocentric video datasets EPIC-KITCHENS~\cite{Damen2018EPICKITCHENS} and EGTEA Gaze+~\cite{li2018eye}.
Results from the leaderboard of the EPIC-KITCHENS action anticipation challenge clearly show our model beats other existing single models.
To summarize, our contributions are summarized as follows:

\begin{itemize}
    \item We propose ImagineRNN that breaks down the long-time action anticipation into a series of step-wise feature predictions of short periods, and then predicts the future action labels upon these imagined features.

    \item We reformulate the future feature prediction problem, and propose to optimize the ImagineRNN by picking the correct future states from lots of distractors, which essentially learns the change of future features compared to the traditional regression loss functions.    

    \item We further replace the ImagineRNN's target by predicting the difference between adjacent frames, which helps the model focus on the feature change along time, leading to better anticipation performance. Experiments with different architectures validate the effectiveness of this change.

\end{itemize}

\section{Related Work}

\subsection{Video Understanding and  Action Recognition}
Deep learning methods have achieved promising performance on the video classification task. Simonyan~\etal~\cite{simonyan2014two}  proposed Two-Stream to utilize both RGB frames and optical flow as the 2D CNN input to modeling appearance and motion, respectively. Temporal Segment Networks (TSN) \cite{TSN2016ECCV} extended the two-stream CNN by extracting features from multiple temporal segments. Tran~\etal~\cite{tran2015learning} proposed a 3D CNN to learn the spatial-temporal information. Moreover, Recurrent Neural Networks (RNNs) are also effective in temporal modeling and have been found useful for video classification in \cite{abu2016youtube,zhu2017bidirectional}. 
More recently, some researchers study the egocentric action recognition problem~\cite{kazakos2019TBN, Sudhakaran_2019_CVPR,sudhakaran2018attention,wang2020symbiotic}.
Sudhakaran~\etal~\cite{Sudhakaran_2019_CVPR} proposed a Long Short-Term Attention model to focus on features from relevant spatial parts. 
Wang~\etal~\cite{wang2020symbiotic} proposed a Symbiotic Attention mechanism to enable the communications between motion features and object features in egocentric videos.
Our method builds on these methods and uses TSN as a base framework to train CNNs for action recognition.

\subsection{Early Action Recognition}
The early action recognition task~\cite{aliakbarian2017encouraging,becattini2017done,de2016online,chen2018part} is to recognize the ongoing action as early as possible from partial observations. 
In this task, the model is only allowed to observe a part of the action videos, and predict the action based on the video segment~\cite{de2018modeling,ma2016learning}.
This task is closed to our target, the action anticipating task.
Differently from these works, in the egocentric anticipating task, the action should be recognized before it starts, so we cannot partially observe the action frames at the time of prediction.

\subsection{Action Anticipation}
Predicting the near future has been widely studied recently~\cite{li2018deep,liang2019peeking,wang2018eidetic,YanBin2020}.
Action anticipation is to predict an action \emph{before} it occurs~\cite{gao2017red,furnari2018Leveraging,8822593}. Previous works investigated different forms of action and activity anticipation~\cite{abu2018will,felsen2017will,Furnari2017,furnari2019rulstm,mahmud2017joint,zeng2017visual,rhinehart2017first,7576681}. 
We share a similar idea with past works and use the recurrent neural networks to summarize the past observations~\cite{abu2018will,gao2017red}. 
Very recently, RULSTM~\cite{furnari2019rulstm} consists of two LSTMs to anticipate actions from egocentric video, where one LSTM is used to summarize the past, and the other is used to predict future actions based on the past future directly. 
Miech~\etal~\cite{miech2019leveraging} proposed to directly anticipate future action based on the combination of past visual inputs and past action recognition results.
Concurrent to us, Sener~\etal~\cite{sener2020temporal} propose a multi-scale temporal aggregate method for action anticipation by relating recent to long-range observations. It computes recent and spanning representations pooled from snippets that are related via coupled attention mechanisms. The experiments shows great advantages brought by ensembles of multiple scales. 

There are also some other interesting researches for the anticipation task.
In \cite{abu2018will}, the authors study the problem of anticipating a sequence of activities within time horizons of up to 5 minutes, in contrast to other works that anticipate the next action within several seconds.
\cite{nagarajan2020ego} studies anticipating future actions in the long-time period. Given an untrimmed video containing a long composite activity, the proposed topological affordance graphs could predict the actions that will likely occur in the future to complete it. 
Ego-OMG~\cite{dessalene2020egocentric} proposes to structure the long video clips into a discrete set of states, where each state represents the objects presently in contact or anticipated to soon be in contact.

Given past observation, it might have many possible future actions due to future uncertainty.
Future uncertainty (alternative future) is important in the action anticipation task. 
Furnari~\etal~\cite{furnari2018Leveraging} study how to explicitly incorporate the uncertainty in the loss functions.
Canuto~\etal~\cite{santos2019action} propose to minimize the model uncertainty instead of maximizing its class probabilities, which could be used as the online decision-making criterion for action anticipation.
In~\cite{abu2019uncertainty}, both an action model and a length model are trained to capture the uncertainty of future activities. 
In this paper, we do not explicitly model the future uncertainty in our method. Given existing video data, we only optimize the model to predict the exact next future action that happens in the video. It is a limitation of our method. We hope to handle uncertainty in our future works.

\begin{figure*}[ht!]
    \begin{center}
        \includegraphics[width=0.85\linewidth]{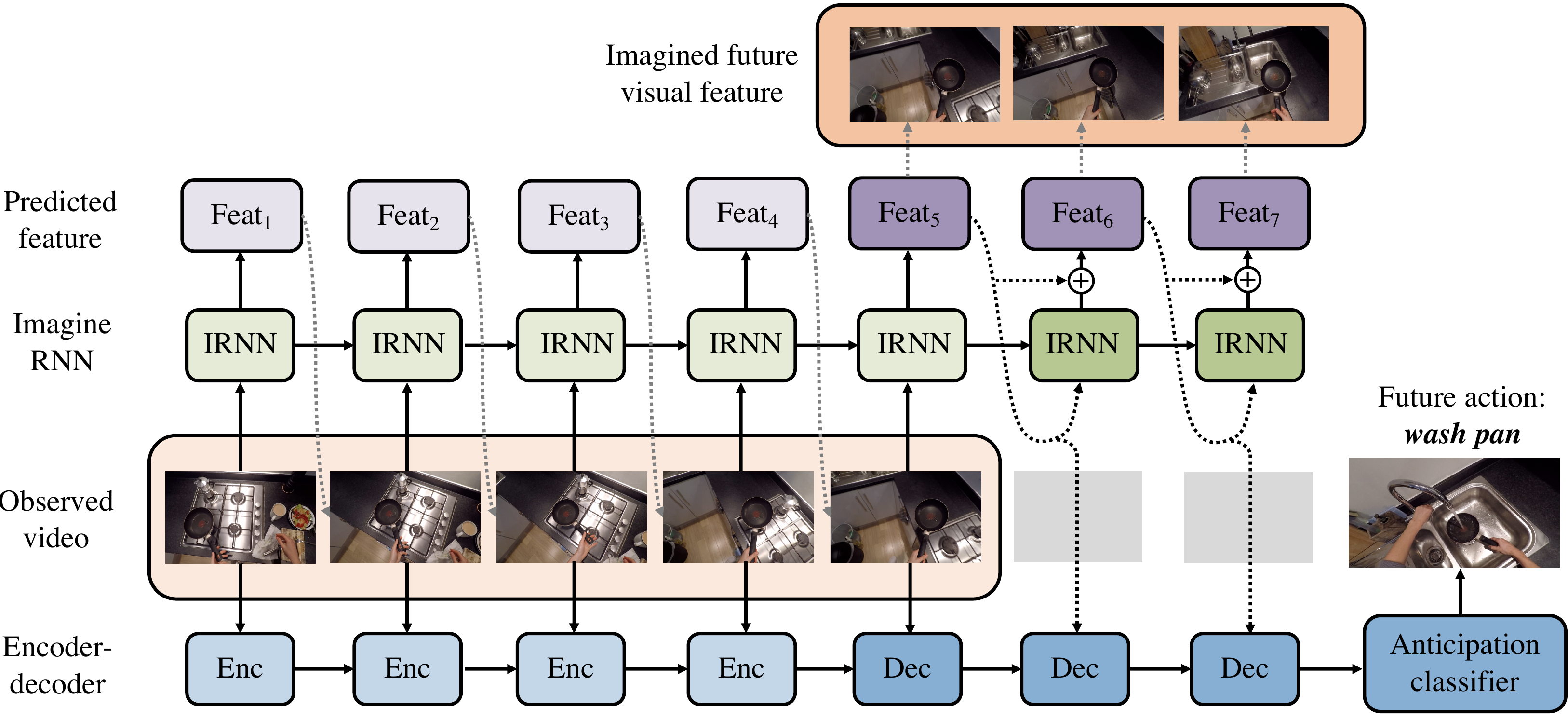}
    \end{center}
    \vspace{-3mm}
    \caption{The framework of our method. ImagineRNN predicts the next visual representation based on past observations in a step-wise manner. The imaginary features are input to the decoder to improve the anticipation performance.
    We propose to better optimize the ImagineRNN with the contrastive learning task. We further improve the ImagineRNN by forecasting the features difference between frames, instead of generating the entire frame representations.
    }
    \label{fig:overview}
\end{figure*}

Some recent works~\cite{gammulle2019predicting,rodriguez2018action} propose to predict the optical flow frames or dynamic image in the future, which has a similar motivation with our designed residual anticipation, \ie, predicting low-entropy signals (the frame-feature difference)
However, different from~\cite{gammulle2019predicting,rodriguez2018action} that predict the entire optical flow frames or dynamic image, we only predict the feature changes between adjacent frames, which avoids wasting model capacity on resolving the stochasticity of frame changes due to camera motion in egocentric videos. 

\subsection{Contrastive Learning}
Contrastive learning aims at optimizing models by distinguishing similar and dissimilar data pairs.
Recent works~\cite{chen2020simple,han2020memory,oord2018representation} proposed to utilize contrastive learning for self-supervised learning. 
Contrastive Predictive Coding (CPC)~\cite{oord2018representation} proposed to learn representation by encoding predictions over future observations from the past.
MoCo~\cite{he2020momentum} designed a momentum encoder and maintained a queue of representations to conduct contrastive learning.
SimCLR~\cite{chen2020simple} experiments with different combinations of data augmentation methods for paired samples in contrastive learning.
Very recently, Han~\etal~\cite{han2020memory} proposed to introduce contrastive learning into the action recognition task. The model is optimized by a predictive attention mechanism over the compressed memories that predicts future representations based on recent observation.
Different these methods, we focus on the action anticipation task rather than representation learning. We found the contrastive learning helps to learn the change of future features, which can be used to obtain better intermediate imaginary data in our ImagineRNN framework.

\section{Proposed Approach}
\label{sec:method}

\subsection{Egocentric Action Anticipation}
\label{sec:Preliminary}
\noindent
\textbf{Task definition.}
In the EPIC-Kitchens anticipation challenge~\cite{Damen2018EPICKITCHENS}, the egocentric action anticipation task is defined to predict the future action one second before it happens.
In a more general task definition~\cite{furnari2019rulstm},
the video is input in an on-line fashion, with a short video snippet consumed every $\alpha$ seconds, \ie, the video is divided into segments of length $\alpha$.  For an action occurring at time $\tau_s$, the model should anticipate the action by observing the video frames before $\tau_s-T$. 
In our framework, our model is allowed to observe the video segment of length $(l-T)$ starting at time $(\tau_s-l)$ and ending at time $(\tau_s-T)$.
Following~\cite{furnari2019rulstm}, we use the same task setting and set $l=3.5s$ and $\alpha=0.25s$. We also validate our model under different anticipate time, \ie, $T \in \{1.75s, 1.5s, 1.25s, 1s, 0.75s, 0.5s, 0.25s\}$.  Note that it is more general compared to the task defined in~\cite{Damen2018EPICKITCHENS}, which only validates the model under anticipate time $T=1$. 

\noindent
\textbf{CNN pretraining.}
The input of our model is the frame-level feature provided by the pre-trained TSN model. In action anticipation, the anticipation targets (objects and actions) do not always appear in the input video, making it hard to learn good representations for CNN models in an end-to-end manner.
To avoid over-fitting and make the CNN model more meaningful, we follow~\cite{furnari2019rulstm} and pre-train the TSN model on the action recognition task. Then the pre-trained CNN weights are fixed during the following training on our action anticipation task. 
We pre-process the videos and obtain different modalities features by pre-trained CNN models, \ie, RGB frame features, optical flow frame features, and the object features.

\noindent
\textbf{Encoder.}
We take a Long Short-Term Memory (LSTM)~\cite{hochreiter1997long}  model as the temporal encoder.
At each time step, the encoder takes as input the visual content that is being observed.
Specifically, at each time-step $t$, we use the pre-trained TSN model to get the current frame feature $\mathbf{f}_t$. Then we input the feature $\mathbf{f}_t$ to update the memory. The new encoding hidden state $\mathbf{h}^E_{t+1}$ is obtained by updating the LSTM unit as follows:
\begin{align}
\mathbf{h}^E_{t+1} = \texttt{Encoder} (\mathbf{f}_t, \mathbf{h}^E_{t}),
\label{eq:lstm}
\end{align}
where $\mathbf{h}^E_{t}$ is the hidden state from the previous forward. We initialize the hidden state as zeros.
To save memory and avoid noises, we only input the frames several seconds before the action occurring time $\tau_s$. 
Following~\cite{furnari2019rulstm}, we take the frames from $(\tau_s-4)$s to $(\tau_s-2.5)$s as the input for the encoder.

\noindent
\textbf{Decoder.}
The decoder is an LSTM model that performs anticipation. It takes the observed information extracted from the EncodingRNN as the initial hidden states, and then recurrently takes the last observed frame as input.
Based on the last output of the DecodingRNN, we use a fully-connected layer as the classifier for the action anticipation prediction.

\subsection{Bridging the gap between past and future}
\label{sec:Decomposing}
In the egocentric action anticipation task, it is hard to train a meaningful model due to the clear gap between past observations and future action.
We alleviate this issue by decomposing the long-time prediction into a series of short-term forecasts. Then
we design ImagineRNN to fill in the gap by producing the future visual representation.
In this way, the long-time reasoning is simplified by predicting the action based on past observations and future imaginary data.

Specifically, we break down the $T$ seconds anticipation into several short-term anticipations with each lasting $\alpha$ seconds ($\alpha < T$).
Given the visual feature $\mathbf{f}_t$ at time $t$, the ImagineRNN is designed to generate the future visual feature $\hat{\mathbf{f}}_{t+1}$ by,
\begin{align}
&\mathbf{h}^I_{t+1} = \texttt{ImagineRNN} (\mathbf{f}_t, \mathbf{h}^I_{t}), \\
&\mathbf{\hat{f}}_{t+1} = \phi(\mathbf{h}^I_{t+1}),
\label{eq:reconstruct}
\end{align}
where $\mathbf{h}^I_{t}$ is the hidden state of ImagineRNN at time step $t$. $\phi(\cdot)$ is a transformation layer that maps the hidden state space to the visual feature space.
The generated visual feature $\hat{\mathbf{f}}_{t+1}$ is supposed to fill in the gap between the past and future. 
In the framework, we input the output of ImagineRNN to the decoder to predict future action.
Thus the prediction of ImagineRNN should be consistent with the ground truth visual content. Next, we illustrate how we optimize the ImagineRNN model efficiently in the action anticipation framework.

\subsection{Optimization of ImagineRNN}
\label{sec:Optimization}
In egocentric videos, the action states usually change very quickly. Thus the predicted future from ImagineRNN should be substantially different along with the anticipation time. 
The commonly used regression loss functions, such as $l_2$ loss, can hardly optimize the ImagineRNN to perceive the changes of action states.
Differently, we propose a more effective optimization for the ImagineRNN by introducing the contrastive learning task, where the model is asked to pick the correct future states from lots of distractors. We use Noise Contrastive Estimation (NCE)~\cite{gutmann2010noise} to encourage the predicted future feature $\hat{\mathbf{f}}_{t+1}$ to be close to the ground truth future state $\hat{\mathbf{f}}_{t+1}$. Compared to the regression losses, NCE does not require to resolve the low-level stochasticity strictly.
Specifically, for the imagined future feature $\hatf_t$ at time $t$, the only positive target is the ground truth feature $\f_t$. We then build a set of candidates as distractors for the ground truth feature $\f_t$ at time $t$.

\textbf{Distractors.}
 The distractors contain easy negatives and hard negatives. 
The easy negatives contain the frame features from the other videos instead of the target video.
We use the frame-level features from the other videos in the same mini-batch as the easy negatives for simplicity in the calculation. 
These candidates are easy to distinguish since these frames usually look different from the current video.

The hard negatives contain the frames from the same video but at different time steps, $\f_t'$ where $t' \ne t$. These candidates are hard to distinguish since they are very close to the ground truth frame feature $\f_t$. 
Distinguishing the hard negatives encourages ImagineRNN to generate essential intermediate features and capture the change of a series of future states.

\textbf{Contrastive Learning.}
With the positive targets and these distractors, we can take the contrastive learning as a proxy task for better optimizing the ImagineRNN.
Inspired by recent representation learning work~\cite{wu2018unsupervised,wu2019progressive}, we first calculate the cosine similarity between the predicted feature and the candidates,
$\v^{\texttt{T}}_j \hatf_t$, where $\v_j$ denotes the $j$-th distractors. 
Here we enforce all vectors to be L2-normalized feature embeddings, \ie, $||\v_j||=1$, $||\hatf_t||=1$, and $||\f_t||=1$.
Thus we have the following objective function at the time step $t$,
\begin{align}\label{eq:nce}
  \mathcal{L}_c = - \log \frac{\exp({\f_t}^{\texttt{T}} \hatf_t/\tau)}{\sum_{j} \exp({\v_j}^\texttt{T} \hatf_t/\tau) + \exp({\f_t}^{\texttt{T}}\hatf_t/\tau)},
\end{align}
\noindent
where $\tau$ is a temperature parameter that controls the concentration level of the distribution. Higher $\tau$ leads to a softer probability distribution. We set $\tau=0.2$ in our experiments.

With Eqn.~\eqref{eq:nce}, we optimize the ImagineRNN with a cross-entropy loss (negative log-likelihood), instead of the commonly used regression loss functions.
During optimization, the loss function encourages the predicted feature $\hatf_t$ to be close to ground truth target $\f_t$, and also push the predicted feature $\hatf_t$ to be distinct from these distractors.
Thus the trained ImagineRNN could catch the change of action states at different times, which is essential in action anticipation.

2) The \textit{future intention}.
In addition, following~\cite{shi2018action,gammulle2019predicting,rodriguez2018action}, we also take the future intentions as additional supervision.
The future intention is the purpose of the currently observed actions (the next future action), which explains the visual changes that happen during the unseen temporal region $T$.
The intuition behinds it is that the generated visual representation should also benefit the anticipation task. Specifically, we input the generated visual feature to the decoder for several time steps during the anticipation time period.
The decoder's last hidden state is further input to the action classifier for recognizing the future action.
Then we use the Cross-Entropy loss on the final action anticipation to optimize the ImagineRNN.
Denote the Cross-Entropy loss of the classifier as $\mathcal{L}_f$, the final loss is the sum of the two losses,
\begin{align}
\mathcal{L} = \mathcal{L}_c + \mathcal{L}_f.
\label{eq:combine}
\end{align}

\subsection{Forecasting the difference between frames}
\label{sec:Diff}
However, the visual features of adjacent frames would be close since the backgrounds in frames are the same.
Directly predicting the visual feature of future frames might waste model capacity in generating the unchanged background information. 
In addition, ImagineRNN might not essentially learn the change during future frames. 
Thus we propose to improve ImagineRNN by explicitly force it to predict the feature difference of adjacent frames, instead of the entire frame feature.

Specifically, we optimize ImagineRNN by learning to produce the difference between the current visual feature and the next one. The output of ImagineRNN is to forecast future \textit{changes} of the visual feature given the current observation.
Thus we change Eqn.~\eqref{eq:reconstruct} to be,
\begin{align}
\label{eq:diff}
\mathbf{\hat{f}}_{t+1} = \phi(\mathbf{h}^I_{t+1}) + \mathbf{f}_t.
\end{align}

Since we are designed to predict a series of intermediate frame features before anticipating the future action, we repeatedly use Eqn.~\eqref{eq:diff} to generate a series of future frame features in an auto-regressive way.
Suppose frame $t$ to be the last observed frame, we can obtain the imagined feature $\mathbf{\hat{f}}_{t+n}$ of future frame $t+n$ by,
\begin{align}
\label{eq:diff2}
\mathbf{\hat{f}}_{t+n} = \phi(\mathbf{h}^I_{t+n}) + \phi(\mathbf{h}^I_{t+n-1}) + ... + \phi(\mathbf{h}^I_{t+1}) + \mathbf{f}_t.
\end{align}

As can be seen in the above equation, predicting the difference sets up a shortcut connection between step-wise reconstructions, which helps ease the optimization of ImagineRNN and avoids noise accumulation during the auto-regressive future feature generation in testing. In addition,
Predicting the frame difference promotes the model to focus on the change of intermediate features, which might be the core of future action anticipation.

\section{Experiments}
We first discuss the experimental setups and then compare our method with the state-of-the-art methods on two large-scale egocentric action datasets, EPIC-Kitchens and EGTEA Gaze+.
Ablation studies and qualitative results are provided to show the effectiveness of our method.

\begin{table*}[]
\centering
\small
\caption{Egocentric action anticipation results on the \textbf{Seen (S1)} test set of the EPIC-KITCHENS Action Anticipation Challenge~\cite{Damen2018EPICKITCHENS} with anticipation time $T=1$ second. 
All values are reported as percentage (\%).}
{
\begin{tabular}{l||c|c||c|c||c|c}
\Xhline{0.8pt}
\multirow{2}{*}{Methods} & \multicolumn{2}{c||}{Verb}      & \multicolumn{2}{c||}{Noun}      & \multicolumn{2}{c}{Action}    \\ \cline{2-7} 
                         & Top-1 Acc & Top-5 Acc & Top-1 Acc & Top-5 Acc  & Top-1 Acc & Top-5 Acc  \\ \hline
2SCNN~\cite{Damen2018EPICKITCHENS} & 29.76 &76.03 &15.15 &38.56 &04.32 & 15.21 \\ 
ATSN~\cite{Damen2018EPICKITCHENS} & 31.81 & {76.56} &{16.22}  &{42.15} &06.00& {28.21}\\ 
ED~\cite{gao2017red} &29.35 & 74.49&16.07& 38.83 &08.08 & 18.19 \\
MCE~\cite{furnari2018Leveraging} & 27.92 &73.59  & 16.09 & 39.32 & {10.76}  & 25.28 \\ 
Transitional~\cite{miech2019leveraging} &30.74 &76.21 &16.47 &42.72 &09.74 &25.44\\
RULSTM~\cite{furnari2019rulstm} & 33.04 &79.55 &22.78 &50.95 &14.39 &33.73 \\ \Xhline{0.8pt}
\textbf{Ours }($l_2$ loss) &35.26 & 79.66 &22.57 &52.04 &\textbf{15.07} &34.66 \\ 
\textbf{Ours} (Contrastive) &\textbf{35.44} &\textbf{79.72} & \textbf{22.79} &\textbf{52.09} & 14.66 & \textbf{34.98} \\ 
\Xhline{0.8pt}
\end{tabular}}
\vspace{0.1cm}
\label{tab:test_seen}
\end{table*}

\begin{table*}[]
\centering
\small
\caption{Egocentric action anticipation results on the \textbf{Unseen (S2)} test set of the EPIC-KITCHENS Action Anticipation Challenge~\cite{Damen2018EPICKITCHENS} with anticipation time $T=1$ second. 
All values are reported as percentage (\%).}
{
\begin{tabular}{l||c|c||c|c||c|c}
\Xhline{0.8pt}
\multirow{2}{*}{Methods} & \multicolumn{2}{c||}{Verb}      & \multicolumn{2}{c||}{Noun}      & \multicolumn{2}{c}{Action}    \\ \cline{2-7} 
                         & Top-1 Acc & Top-5 Acc & Top-1 Acc & Top-5 Acc  & Top-1 Acc & Top-5 Acc  \\ \hline
2SCNN~\cite{Damen2018EPICKITCHENS}& 25.23 &68.66 & 09.97 & 27.38 & 02.29 & 09.35 \\ 
ATSN~\cite{Damen2018EPICKITCHENS} & {25.30} & 68.32 & {10.41} &{29.50} & 02.39  & 06.63 \\
ED~\cite{gao2017red} &22.52 & 62.65&07.81& 21.42 &02.65 & 07.57 \\
MCE~\cite{furnari2018Leveraging} & 21.27  & 63.33 & 09.90  & 25.50 & {05.57} & {15.71} \\
Transitional~\cite{miech2019leveraging} &28.37 &69.96 &12.43 &32.20 &07.24 &19.29\\
RULSTM~\cite{furnari2019rulstm} & 27.01 & 69.55 & 15.19 & 34.38 &08.16 & 21.10 \\ \Xhline{0.8pt}
\textbf{Ours }($l_2$ loss) &27.35  &69.78 &15.36  &35.34  &08.54  &20.79  \\ 
\textbf{Ours }(Contrastive) &\textbf{29.33}  &\textbf{70.67}  & \textbf{15.50}  &\textbf{35.78}  & \textbf{09.25}  & \textbf{22.19}  \\
 \Xhline{0.8pt}

\end{tabular}}
\label{tab:test_unseen}
\end{table*}

\subsection{Experimental Settings} \label{section:dataset}
\noindent
\textbf{Datasets.}
We perform experiments on two large-scale datasets of egocentric videos: EPIC-Kitchens~\cite{Damen2018EPICKITCHENS} and EGTEA Gaze+~\cite{li2018eye}. 
\textbf{EPIC-Kitchens }is the largest dataset in first-person vision so far. It consists of 55 hours of recordings capturing all daily activities in the kitchens.
The activities performed are non-scripted, which makes the dataset very challenging and close to real-world data. 
This dataset is densely annotated with timestamps for each action so that it is ready for the action anticipation task.
Actions in the EPIC-Kitchens dataset is annotated in the format of $(\texttt{verb}, \texttt{noun})$ pairs.
The dataset contains $39,596$ action annotations, $125$ verbs, and $352$ nouns. 
We considered all unique $(\texttt{verb}, \texttt{noun})$ pairs in the public training set, thus obtaining $2,513$ unique actions.
We use the same split as~\cite{furnari2019rulstm} and split the public training set of EPIC-Kitchens ($28,472$ action segments) into training ($23,493$ segments) and validation ($4,979$ segments) sets.
\textbf{EGTEA Gaze+} contains $19$ verbs, $51$ nouns and $106$ unique actions.
We report the average performance across the three official splits provided by the authors of the dataset.

\noindent
\textbf{Evaluation Metrics.}
Following~\cite{furnari2019rulstm}, we use the Top-k accuracy to evaluate our method.
Under this evaluation metric, the prediction is deemed to be correct if the ground truth action falls in the top-k predictions. 
This metric is appropriate due to the uncertainty of future predictions ~\cite{furnari2018Leveraging,koppula2016anticipating}.
Many possible actions can follow an observation.
We use the Top-5 accuracy as a class-agnostic measure. We also report Mean Top-5 Recall~\cite{furnari2018Leveraging} as a class aware metric. 
Top-5 recall for a given class $c$ is defined as the fraction of samples of ground truth class $c$ for which the class $c$ is in the list of the top-5 anticipated actions.
Mean Top-5 Recall averages Top-5 recall values over classes.
In~\cite{furnari2018Leveraging}, Top-5 Recall is averaged over the provided list of many-shot verbs, nouns, and actions. 
Performances are evaluated for verb, noun, and action predictions. 
Following~\cite{furnari2019rulstm}, in training the only targets are the action labels, and our model is optimized to predict the action prediction. In the testing, we obtain the predictions for verb and noun by the marginalization on action predictions.

\noindent
\textbf{Implementation Details.}
We use Pytorch~\cite{paszke2017automatic} to implement our framework.
For the pre-trained action recognition model, we use a BNInception CNN~\cite{ioffe2015batch} with the TSN framework to train the action recognition model.
After pre-training, we resize the frame to  $456 \times 256$ pixels and input it to the CNN model.
The output ($1024$-dimensional vectors) of the last global average pooling layer is used as the frame-level feature.
The encoder, decoder, and the ImagineRNN are all single-layer LSTMs with $1024$ hidden units. 
We use Stochastic Gradient Descent (SGD) to train the framework with a mini-batch size of $128$ and a learning rate of $0.01$ and momentum equal to $0.9$.
We train 100 epochs and apply early stopping at each training stage the same as~\cite{furnari2019rulstm}. This is done by choosing the intermediate and final models' iterations, which obtain the best Top-5 action anticipation accuracy for the anticipation time $T=1s$ on the validation set. 
Following~\cite{furnari2019rulstm}, we use the RGB frames, optical flow frames, and the object detection features as input for our model. We first train the model with each modality individually and then obtain the final prediction by a late fusion of the three models' predictions.
In the following experiments, for fair comparisons with RULSTM, our model takes all the three modalities as input if not specified.

\subsection{Comparison to the state-of-the-art methods}
\noindent
\textbf{Compared Methods\hspace{0.5em}}
We compare our method with state-of-the-art action anticipation methods:
Deep Multimodal Regressor (DMR)~\cite{vondrick2016anticipating},
Anticipation Temporal Segment Network (ATSN) of~\cite{Damen2018EPICKITCHENS}, Anticipation Temporal Segment Network trained with verb-noun Marginal Cross Entropy Loss (MCE)~\cite{furnari2018Leveraging}, and the Encoder-Decoder LSTM (ED) introduced in~\cite{gao2017red}. 
We also compare with the early action recognition methods to the problem of egocentric action anticipation: Feedback Network LSTM (FN)~\cite{de2018modeling}, and an LSTM trained using the Exponential Anticipation Loss~\cite{jain2016recurrent} (EL). 
To compare with state-of-the-art action anticipation methods, we reproduced a vanilla version of Feature Mapping RNN~\cite{shi2018action} without the kernalised RBF. 
For a fair comparison, we first train models with the three input modalities, \ie, RGB features, optical flow features, and the object features. Then we obtain the final prediction by a late fusion of the three models. 
Very recently,  RULSTM~\cite{furnari2019rulstm} is proposed by combining two LSTM to anticipate actions from egocentric video, where one LSTM is used to summarize the past, and the other is used to predict future actions based on the past future. 
We compare our method under both the standard anticipation setting (anticipation time $T=1s$) and a more general anticipation setting (with variant anticipation time).

\begin{table}
    \centering
    \caption{Action anticipation results on the EPIC-KITCHENS validation set under different anticipation time $T$. The performance is measured by the top-5 accuracy of action anticipation. }
    \small
        \begin{tabular}{p{1.6cm}|cccccc}
            \hline
            \multirow{2}{*}{Methods} & \multicolumn{6}{c}{Top-5 Action Accuracy @ different $T$} \\   \cline{2-7}  %
             & $1.5$ & $1.25$ & $1.0$& $0.75$&$0.5$&$0.25$ \\ \hline
            DMR~\cite{vondrick2016anticipating}             & /                   & /              & 16.9          & /              & /              & /                        \\
            ATSN~\cite{Damen2018EPICKITCHENS}           & /                          & /              & 16.3          & /              & /              & /                \\
            MCE~\cite{furnari2018Leveraging}           & /                        & /              & 26.1          & /              & /              & /            \\
            FMRNN~\cite{shi2018action} & /              & /                           &32.7           & /              & /              & /            \\
            ED~\cite{gao2017red}                           & 23.2          & 24.8          & 25.8          & 26.7          & 27.7          & 29.7         \\
            FN~\cite{de2018modeling}                      & 24.7         & 25.7          & 26.3          & 26.9          & 27.9          & 29.0        \\
            EL~\cite{jain2016recurrent}                   & 26.4          & 27.4          & 28.6          & 30.3          & 31.5          & {33.6}   \\ 
            RULSTM~\cite{furnari2019rulstm}                     & 32.2          & 33.4          & 35.3          & 36.3          & 37.3          & {39.0}   \\ \hline
            \textbf{Ours}                & \textbf{32.5}          &\textbf{33.6}           & \textbf{35.6}         & \textbf{36.7}         &\textbf{38.5}         & \textbf{39.4}   \\ \hline
        \end{tabular}
    \label{tab:different_t}
\end{table}

\noindent
\textbf{Results on the EPIC-KITCHENS test server.} 
We compare our method with the state-of-the-art methods on the test server of EPIC-KITCHENS. 
Table~\ref{tab:test_seen} and Table~\ref{tab:test_unseen} report results obtained from the official EPIC-KITCHENS action anticipation challenge submission server. 
The official test server computes the performances on two test sets, \ie, the ``seen'' test, which includes the same scenes appearing in the training set ({S1}) and the ``unseen'' test set ({S2}), with kitchens not appearing in the training set.
On both test sets, our method outperforms all previously reported results under all metrics. 
On the S1 (seen) test set (Table~\ref{tab:test_seen}), our method outperforms the previous method RULSTM by 1.25\% on the Top-5 Action accuracy.
On the S2 (unseen) test set (Table~\ref{tab:test_unseen}) where the videos are captured in new environments, our method significantly improves RULSTM in all metrics on Verb, Noun, and Action prediction. 
Note that we use the same input features with RULSTM, thus the comparison with RULSTM is a fair comparison, and the performance improvements over RULSTM are all from our algorithm instead of better features.
These results demonstrate our method is better at anticipating the future action.

\noindent
\textbf{Results with Different Anticipation Time $T$.} 
Our method can also be used to predict future action under different anticipation time.
Since each time step $\alpha$ in our method is 0.25s, we can evaluate the future anticipation every 0.25s.
We compare our method with the state-of-the-art methods under different anticipation time $T \in$ $\{$2s, 1.75s, 1.5s, 1.25s, 1s, 0.75s, 0.5s, 0.25s $\}$.
The results are shown in Table~\ref{tab:different_t}.
Note that some methods~\cite{Damen2018EPICKITCHENS,furnari2018Leveraging,vondrick2016anticipating} can anticipate actions only at a fixed anticipation time. 
We found the proposed method always outperforms the strong competitor RULSTM~\cite{furnari2019rulstm} under all anticipation time $T$.
Note that the results are reported on the validation set, where the models are selected by choosing the best validation performance, as used by RULSTM~\cite{furnari2019rulstm}.
As indicated in~\cite{furnari2019rulstm}, the results on the test server are more important in evaluating compared to the validation results.

\noindent
\textbf{Results on the EGTEA Gaze+ dataset.} 
We also conduct experiments on the  EGTEA Gaze+ dataset.
Table~\ref{tab:anticipation_egtea} reports Top-5 action accuracy scores on EGTEA Gaze+ under different anticipation times. 
We use the same input modalities as RULSTM.
Our method outperforms the compared methods under different anticipation time $T$. We also found the relative improvement is smaller on the EGTEA Gaze+ dataset compared to that on the EPIC-KITCHENS dataset. It might be because the EGTEA Gaze+ is relatively small in scale. It only consists $106$ actions, which is far less than the $2,513$ actions in EPIC-KITCHENS. Thus the anticipation on the EPIC-KITCHENS dataset is more challenging.

\begin{table}[t]
    \caption{Comparison of the anticipated action accuracies with different modalities on the validation set.  ``Obj'' indicates the object features. ``w/o intention'' is the model optimized without the future intention (Eqn.~\eqref{eq:combine}).
    ``w/o diff'' indicates the model without forecasting the different.}
    \small
    \centering
    \begin{tabular}{l|l|cc}
        \hline
        Modality & Method & Top-1 Acc & Top-5 Acc  \\ \hline
        
        \multirow{4}{*}{RGB} 
        & RULSTM~\cite{furnari2019rulstm}& 13.05 & 30.83 \\
        & Ours w/o intention &13.23 &31.39 \\  
        & Ours w/o diff & {12.97} & {30.61}  \\
        & Ours & \textbf{13.68} & \textbf{31.58} \\ \hline
        
        \multirow{4}{*}{Flow} 
        & RULSTM~\cite{furnari2019rulstm}& 08.77 & 21.42 \\
        & Ours w/o intention &08.81 &21.89 \\
        & Ours w/o diff & {08.51} & {21.68}  \\
        & Ours & \textbf{09.23} & \textbf{22.06} \\ \hline
        
        \multirow{4}{*}{Obj} 
        & RULSTM~\cite{furnari2019rulstm}& 10.04 & 29.89 \\
        & Ours w/o intention & \textbf{10.76} &30.05 \\
        & Ours w/o diff & {10.62} & {30.12}  \\
        & Ours  & 10.72 & \textbf{30.27} \\ \hline
        
        \multirow{4}{*}{Fusion} 
        & RULSTM~\cite{furnari2019rulstm}& 15.00 & 35.24 \\
        & Ours w/o intention & 15.04 &35.17 \\
        & Ours w/o diff & {14.91} & {34.98}  \\
        & Ours  & \textbf{15.23} & \textbf{35.38} \\ 
        
        \hline
    \end{tabular}
    \label{tab:diff}
\end{table} %

\begin{figure}[t!]
    \begin{center}
        \includegraphics[width=0.7\linewidth]{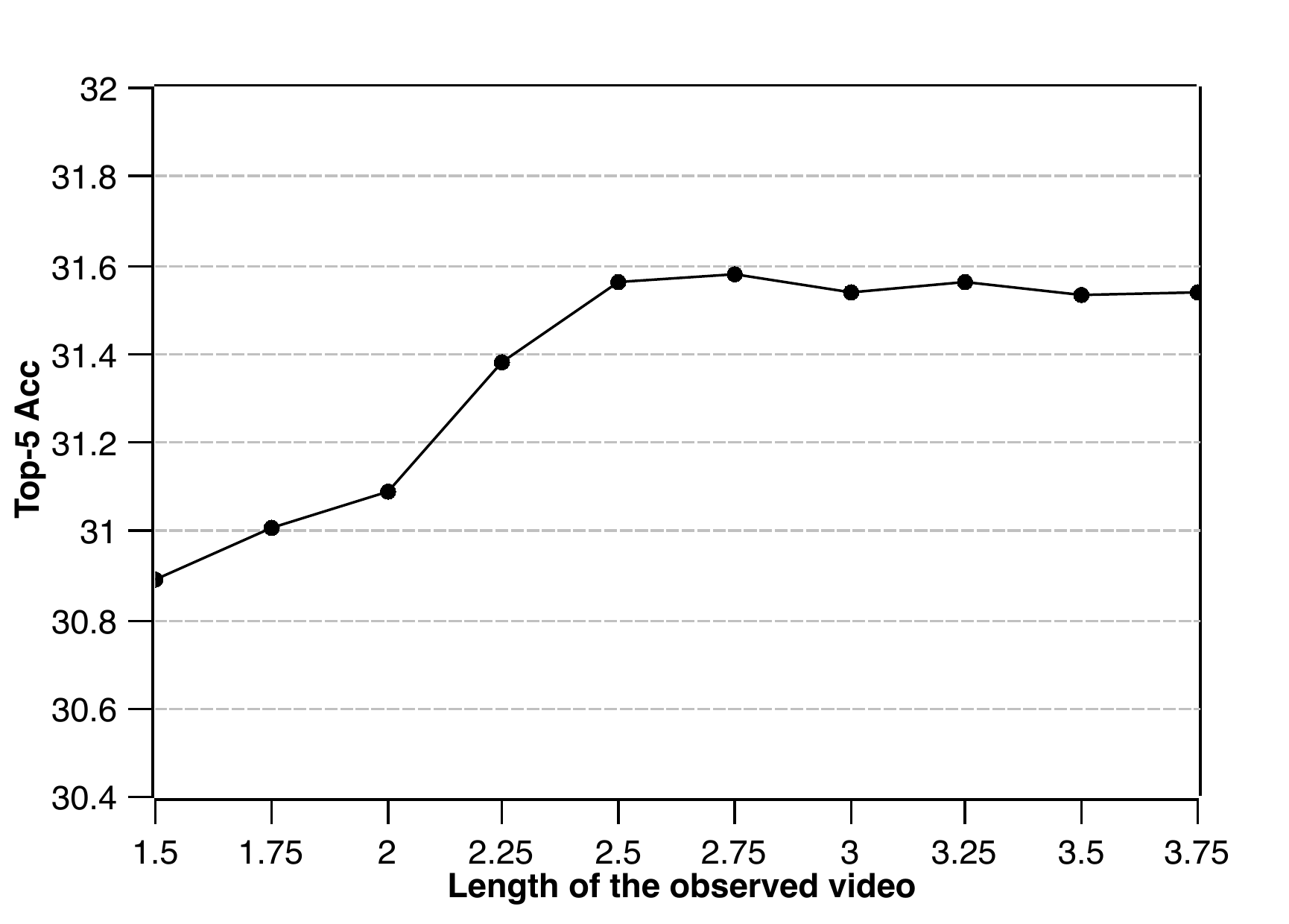}
    \end{center}
    \vspace{-4mm}
    \caption{Top-5 accuracies over different lengths of observed past for the encoder. The results are produced by our method with the RGB modality input. Note the anticipation time $T$ is 1s for all experiments. }
\label{fig:different_length}
\end{figure}

\begin{figure*}[t!]
    \begin{center}
        \includegraphics[width=0.8 \linewidth]{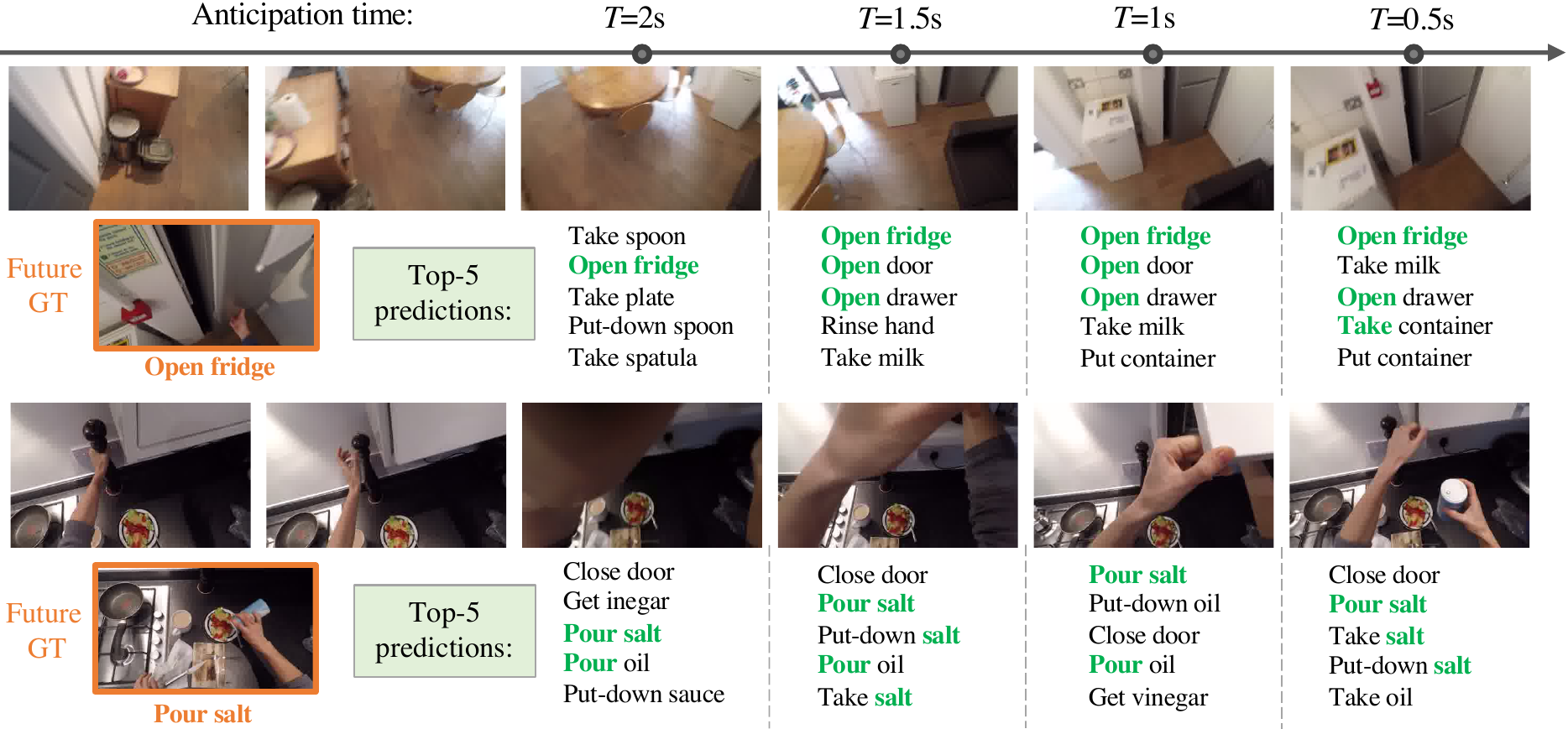}
    \end{center}
    \vspace{-4mm}
    \caption{Qualitative results with anticipation time $T=2s, 1.5s, 1s, 0.5s$. From left to right, the observations are getting closer to future action. We list the Top-5 action predictions of our model at each anticipation time.
    Orange indicates the ground truth, and green means our prediction matches the ground truth.}
\label{fig:visualization}
\end{figure*}

\subsection{Ablation Studies}
We conduct ablation studies to evaluate the effectiveness of the two components of our method.

\noindent
\textbf{Effectiveness of ImagineRNN.} 
Without our proposed ImagineRNN, the model is the baseline RULSTM. From Table~\ref{tab:test_seen} and Table~\ref{tab:test_unseen}, we can see the results of our baseline model only achieve 33.73\% in Top-5 accuracy on the seen (S1) test set and 21.10\% on the unseen (S2) test set. 
By adding our ImagineRNN to the framework, we observed a clear performance improvement on both test sets.

\begin{table}
    \caption{Anticipation results on the EGTEA Gaze+ dataset.}
    \setlength{\tabcolsep}{8pt}
    \begin{adjustbox}{width=\linewidth,center}
        \begin{tabular}{p{2cm}|cccc}
            \hline
            \multicolumn{1}{c}{} & \multicolumn{4}{c}{Top-5 Action Accuracy\% @ different $T$}   \\  \hline %
            \multicolumn{1}{c}{} & $1.0$ & $0.75$ & $0.5$ & $0.25$ \\ \hline
            DMR~\cite{vondrick2016anticipating}  & 55.70  &/  &/  &/                    \\
            ATSN~\cite{Damen2018EPICKITCHENS}  & 40.53    & /              & /              & /                                         \\
            MCE~\cite{furnari2018Leveraging}  & 56.29                & /              & /              & /                           \\
            ED~\cite{gao2017red}                       & 50.22       & 51.86        & 49.99          & 49.17                \\
            FN~\cite{de2018modeling}                   & 60.12          & 62.03          & 63.96          & 66.45                \\
            RL~\cite{ma2016learning}                 & 62.56          & 64.65          & 67.35          & 70.42                 \\
            EL~\cite{jain2016recurrent}        & {64.62} & {66.89} & {69.60} & {72.38}   \\
            {RULSTM}~\cite{furnari2019rulstm}  &66.40  & {68.41} & {71.84} & {74.28}   \\ \hline
            Ours & \textbf{{66.71}} & \textbf{{68.54}} & \textbf{72.32} & \textbf{74.59} \\
            \hline     
        \end{tabular}
    \end{adjustbox}
    \label{tab:anticipation_egtea}
\end{table}

\noindent
\textbf{Effectiveness of Contrastive Learning.} 
The common used optimization for the ImagineRNN is the regression loss functions ($l_2$ loss). In Table~\ref{tab:test_seen} and Table~\ref{tab:test_unseen}, we show the comparison of different optimization methods on the test set of EPIC-KITCHENS. 
Ours ($l_2$ loss) indicates our models optimized by the $l_2$ loss, while Ours (Contrastive) is the model optimized in the Contrastive Learning way, \ie, picking the correct one from lots of distractors.
With $l_2$ loss, our model achieves 34.66 on Top-5 Accuracy in the seen test set.
In contrast, with Contrastive Learning, our method achieves 34.98\% on Top-5 action accuracy.
The improvement of Contrastive Learning is more clear in the unseen test set.
With the proposed Contrastive Learning, the action anticipation result on the unseen set shows a 1.40\% (22.19\% versus 20.79\%) improvement on the Top-5 action accuracy.
The significant performance gap shows that contrastive learning is a better way to optimize ImagineRNN. It leads to a better generalisability across the various benchmarks.

\noindent
\textbf{Effectiveness of Forecasting the Difference.} 
In Table~\ref{tab:diff}, we show the comparison of results with and without forecasting the difference.
We conduct ablation studies on the RGB input, the flow input, and the fused modalities input. 
The results show a steady improvement by introducing to forecast the difference. Specifically, our method significantly outperforms the one (w/o diff) by 0.9 points on the Top-5 action accuracy on the RGB modality. Similarly, we found our approach also suppresses the model (w/o diff) with optical flow data as inputs. 
These comparison results prove the effectiveness of forecasting the difference instead of directly generating the whole visual feature.
We also validate the effectiveness of forecasting the difference with other architectures. We replace the basic architectures of our ImagineRNN and the encoder-decoder by Gated Recurrent Unit (GRU), instead of the previously used LSTM. The results are shown in Table~\ref{tab:gru}. It can be seen that our predicting the feature difference of adjacent frames still performs better with the GRU-based architecture.

\noindent
\textbf{Effectiveness of Future Intention.} 
We also show the comparison of results with and without the future intention optimization Eqn.~\eqref{eq:combine} in Table~\ref{tab:diff}.
The ablation studies show a small improvement brought by future intention.
Specifically, our final model outperforms the one without future intension on the RGB and flow modalities by about 0.4\% in Top-1 accuracy and 0.2\% in Top-5 accuracy.

\noindent
\textbf{Ablation studies over different lengths of the past.} 
We show the results over different lengths of observed past in Figure~\ref{fig:different_length}. Note the anticipation time $T$ is 1s for all experiments.
It can be seen from the figure that the performance is relatively low if the encoder period is too short (\ie, less than 2.25 seconds). 
As the encoding period gets longer, we found the performance gets steady.
Inputting more observed frames did not lead to further performance improvement if the encoder period is longer than 2.5 seconds.
The reason might be that actions usually change quickly in egocentric videos. Too early frames do not have strong correlations with the future action.

\begin{table}[t!]
\centering
\caption{Comparison of different optimizations on the {Unseen (S2)} test set of the EPIC-KITCHENS Action Anticipation Challenge.
``Con.'' indicates the contrastive learning loss. 
``Adv.'' indicates the adversarial loss used in GAN~\cite{gammulle2019predicting}.
}
\setlength{\tabcolsep}{4.8pt}
\label{tab:different_optimization}
\begin{tabular}{l|cc|cc|cc}
\Xhline{0.8pt}
\multirow{2}{*}{Methods} & \multicolumn{2}{c|}{Verb}      & \multicolumn{2}{c|}{Noun}      & \multicolumn{2}{c}{Action}    \\ \cline{2-7} 
                         & Top-1 & Top-5 & Top-1 & Top-5  & Top-1 & Top-5  \\ \hline
$l_2$ &27.4  &69.8 &15.4  &35.3  &8.5  &20.8  \\ 
Con. + $l_2$ + Adv. &27.9      &70.3     &14.3      &34.7      &8.5       &20.7      \\ 
Con. + $l_2$ &28.4  &70.0  &15.1   &34.9  &9.0   &21.1   \\ 
Con. &\textbf{29.3}  &\textbf{70.7}  & \textbf{15.5}  &\textbf{35.8}  & \textbf{9.3}  & \textbf{22.2}  \\

\Xhline{0.8pt}

\end{tabular}
\end{table}

\begin{table}[t!]
\centering
\caption{Ablation studies of predicting the feature difference between adjacent frames with GRU-based architecture on the EPIC-KITCHENS Action Anticipation validation set.
}
\setlength{\tabcolsep}{4.6pt}
\label{tab:gru}
\begin{tabular}{l|cc|cc|cc}
\Xhline{0.8pt}
\multirow{2}{*}{Methods} & \multicolumn{2}{c|}{Verb}      & \multicolumn{2}{c|}{Noun}      & \multicolumn{2}{c}{Action}    \\ \cline{2-7} 
                         & Top-1 & Top-5 & Top-1 & Top-5  & Top-1 & Top-5  \\ \hline
{Ours} (GRU w/o diff) &32.9  &78.7 &22.0  &49.2  &13.3  &32.3 \\ 
{Ours }(GRU with diff) &\textbf{33.7}  &\textbf{79.7}  & \textbf{22.7}  &\textbf{50.2}  & \textbf{14.0}  & \textbf{33.2}  \\
\Xhline{0.8pt}

\end{tabular}
\end{table}

\noindent
\textbf{Discussion on different optimization methods.} 
We evaluate the models with different optimization methods on the test set of the EPIC-KITCHENS Action Anticipation Challenge. 
The results are shown in Table~\ref{tab:different_optimization}, where ``Con.'' indicates the contrastive learning loss, and ``Adv.'' is the adversarial loss used in GAN~\cite{gammulle2019predicting}.
It can be seen that a combination of contrastive loss and $l_2$ loss does not outperform the one with the contrastive learning only. 
Besides, we add the adversarial loss in the model training, where the discriminator is a three-layer MLP.
According to the validation results, we set the weight of the adversarial loss to be 0.01 in the overall loss function and the discriminator's learning rate to be $2 \times 10^{-6}$.
As shown in Table~\ref{tab:different_optimization}, the model trained with the combination of the three loss functions performs worst among all candidates. 
Our model trained with contrastive learning performs best among all candidates. The reason might be that contrastive learning helps to learn the change of future features essentially, since it needs to distinguish the positive target from lots of distractors. (frame features at other times).

\subsection{Qualitative Results}
We show some qualitative results of our method in Fig~\ref{fig:visualization}. 
From left to right, the observations are getting closer to future action.
The orange box and ``GT'' indicates the ground truth of the future action.
We list the Top-5 action predictions of our results at the anticipation time $T \in \{2s, 1.5s, 1s, 0.5s\}$.
Green indicates the prediction matches the ground truth.
Taking the first row as an example, the anticipations become more and more accurate as time flows. It is consistent with our motivation that long-time modeling might involve lots of noise. 
It is interesting to see the model always predicts ``Open fridges'' when $T$ is less than 2 seconds probably because the fridge shows up in the observations at $T=1.5s$.
The other action candidates, including ``Take milk'' and ``Open drawer'', are also likely to take place in the near future.

\section{Conclusions and future works}
In this work, we decompose the action anticipation task into a series of future frame feature predictions. We first imagine how the future feature changes and then predict future action based on these imagined representations.
We found that ImagineRNN optimized with contrastive learning is superior than the typical anticipation models.
In addition, we further propose to improve ImagineRNN by predicting the feature difference of adjacent frames instead of the whole frame content. 
It helps promote the model to focus on the change of future states and avoid the noise accumulation during the auto-regressive future feature generation.
Extensive experimental results on different architectures validate the effectiveness of the proposed method.

In conclusion, we found it useful by decomposing action anticipation into lots of intermediate predictions. 
Focusing on the future state transition by contrastive learning and predicting future frames' difference improves the quality of intermediate predictions, leading to better results on the final action anticipation task.

In future works, we might further explore the uncertainty of future in the egocentric action anticipation task, which is a limitation of our current work. It might also be helpful in generating better intermediate future features by considering the hand tracking on the observed frames.

\section{Acknowledgement}
Yu Wu is the recipient of the Google PhD Fellowship 2020. We acknowledge the Google PhD Fellowship Programme and ARC Discovery Project DP200100938 for funding this research.

{
\bibliographystyle{IEEEtran}
\bibliography{egbib}
}

\end{document}